\documentclass[11pt]{article}

\usepackage[final]{acl}

\usepackage{times}
\usepackage{latexsym}
\usepackage{mathtools}

\usepackage[T1]{fontenc}

\usepackage[utf8]{inputenc}

\usepackage{microtype}

\usepackage{inconsolata}

\usepackage{graphicx}

\title{How Should We Model the Probability of a Language?}

\author{Rasul Dent$^1$, Pedro Ortiz Suarez$^2$, Thibault Clérice$^1$,  Benoît Sagot$^1$ \\ $^1$Inria, Paris, \texttt{\{firstname.lastname\}@inria.fr} \\ $^2$Common Crawl Foundation, Paris, \texttt{pedro@commoncrawl.org}}

\begin{document}
\maketitle
\begin{abstract}
Of the over 7,000 languages spoken in the world, commercial language identification (LID) systems only reliably identify a few hundred in written form. Research-grade systems extend this coverage under certain circumstances, but for most languages coverage remains patchy or nonexistent. This position paper argues that this situation is largely self-imposed. In particular, it arises from a persistent framing of LID as decontextualized text classification, which obscures the central role of prior probability estimation and is reinforced by institutional incentives that favor global, fixed-prior models. We argue that improving coverage for tail languages requires rethinking LID as a routing problem and developing principled ways to incorporate environmental cues that make languages locally plausible.
\end{abstract}

\section{Introduction}

To use many Natural Language Processing (NLP) systems, we must first specify the language. The task of inferring this information is known as automatic language identification (LID).  Following substantial progress in the 1990s, \citet{mcnameeLanguageIdentificationSolved2005} famously described  LID for lengthy European-language documents as “a solved problem suitable for undergraduate instruction.” Attention has since shifted to closely-related varieties \citep{aepliFindingsVarDialEvaluation2023}, rare languages \citep{caswellLanguageIDWild2020}, short texts \citep{murthyLanguageIdentificationSmall2006}, and code-switching \citep{burchellCodeSwitchedLanguageIdentification2024}.

Despite improvements on benchmarks, the practical reality for most languages has changed little in the past 20 years. New models are occasionally released for a region or family \citep[e.g.][]{adebaraAfroLIDNeuralLanguage2022}, and some attempt to reach “the next 1000 languages” \citep[e.g.][]{brownSelectingWeightingNGrams2013, kargaranGlotLIDLanguageIdentification2023}. Nonetheless, production-grade LID systems at companies like Meta and Google still focus on only a few hundred widely spoken languages.

This stagnation reflects two core issues. First, in research settings, LID is typically framed  as a decontextualized inference task, in which systems aim to map directly text to a label from a global set. This encourages approaches that perform well on benchmarks but fail when applied to real data. Second, field-level incentive structures reward developing \emph{novel} methods over revisiting basic tools.

We argue that recentering context will be crucial for expanding the \textit{effective} coverage of LID. For ``local languages'' \citep[see][]{birdLocalLanguagesThird2022} in particular, success should be defined relative to local contexts and resource constraints.  This entails rethinking the values that shape research in LID.



\section{The Received Framing of LID}
\label{The Received Framing of LID}
At its core, LID is intended to route content to users who manage different languages. Due to differences between the modalities of text, speech, and sign, LID in each modality is generally approached separately, with text-based LID  typically viewed as the easiest \citep{jauhiainenAutomaticLanguageIdentification2024}. For \citet{rauLanguageIdentificationStatistical1974},  the prototypical LID operator was a data entry clerk filing documents in unfamiliar languages for human patrons. With the explosion of born-digital content, the immediate recipient is now often a computer program.\footnote{However, Rau's  vision still lives on in some institutions, like libraries.}

\subsection{Standard Approaches}
LID is typically approached as a supervised classification problem \citep{jauhiainenAutomaticLanguageIdentification2024}.
This framing builds in two critical assumptions:



\begin{enumerate}
    \item Labels represent a global hypothesis space.
    \item Inference can be performed solely from text.
\end{enumerate}

As in other classification tasks, there are two main modeling approaches. Per-class approaches such as Multinomial Naïve Bayes \citep[e.g.][]{luiLangidPyOfftheshelf2012} fit the data to each class independently, and take the best fitting class as the label.\footnote{Traditionally, this is called generative modeling, but as an anonymous reviewer noted, the term has become ambiguous.} In contrast, discriminative models like fastText \citep{joulinBagTricksEfficient2017, graveLearningWordVectors2018} and CLD3 \citep{alexsalcianuCompactLanguageDetector2023} learn decision boundaries between all classes simultaneously.

Hierarchical architectures also garnered interest. Many models, such as IndicLID \citep{madhaniBhasaAbhijnaanamNativescriptRomanized2023}, are region-specific and  depend on external software to ensure that they are used over the correct language set. Others, especially those based on fastText such as OpenLID \citep{burchellOpenDatasetModel2023} and GlotLID \citep{kargaranGlotLIDLanguageIdentification2023}, take a flat approach and model a wide variety of languages using one combined model. Yet others, such as LIMIT \citep{agarwalLIMITLanguageIdentification2023}, try to learn hierarchical classification schemas directly from model errors.

\subsection{Alternative Framings}
The standard assumptions, and the modeling constraints that come with them, are not unreasonable in many  common scenarios. In particular, they produce strong results on well-written monolingual documents in widely spoken languages \citep{mcnameeLanguageIdentificationSolved2005}. However, their reliability deteriorates in several scenarios. \citet{caswellLanguageIDWild2020} and \citet{kreutzerQualityGlanceAudit2022} showed that the models that perform the best on common benchmarks often struggle  when applied to noisy web data. Similarly, when working with closely-related varieties, or very short texts, it is often impossible to select just one correct label on the basis of input alone. The recent Shared Task on Improving Language Identification for Web Text at the 1st Workshop on
Multilingual Data Quality Signals confirmed these issues remain relevant \citep{suarezCommonLIDReevaluatingStateoftheArt2026}.

In response to such issues, alternate framings have been proposed. For example, \citet{baimukanHierarchicalAggregationDialectal2022a}  show that hierarchical labels are important for fine-grained dialect classification. \citet{bernier-colborneDialectVariantIdentification2023} and \citet{kelegArabicDialectIdentification2023} extend this insight, reconceptualizing the task of dialect identification  as one of multi-label classification. From a different angle, \citet{dentIdentifyingRareLanguages2025} contend that building web corpora for rare languages is more of a mining task than strict classification.

Together, these reframings touch on a much more general issue, but do not completely resolve it. Namely, general-purpose LID models need a label set fixed enough to allow training and flexible enough to handle great variation in inference-time granularity. In Section \ref{Context is Non-Negotiable}, we argue that the diversity of inference-time conditions  has often been overlooked, leading to  structural limitations with significant practical and scientific costs.

\section{Probability Problems}
\label{Context is Non-Negotiable}

Both per-class and discriminative approaches to LID ultimately estimate a conditional probability distribution: given input features $X$,
estimate the probability $P$ of each language $\ell$, and choose the most likely label. Bayesian framing reveal that $P(\ell \mid X)$ actually depends on two terms, the probability of $X$ given $\ell$, and the prior probability of $\ell$ itself. The decision rule can be written in the familiar argmax form:

\begin{equation}
\hat{\ell} = \arg\max_{\ell} \; P(X \mid \ell)\, P(\ell),
\end{equation}

or equivalently in log-space:

\begin{equation}
\hat{\ell} = \arg\max_{\ell} \left[ \log P(X \mid \ell) + \log P(\ell) \right].
\end{equation}

This decomposition highlights a highly non-trivial modeling problem. Namely, what does $P(\ell)$ actually represent?

\subsection{Global Frequency?}
In the classic Bayesian formulation, the answer is simple enough. $P(\ell)$ is the number of texts in a given language over the number of examples in the entire corpus. In principle, this should correspond to a real difference in the frequency of languages, rather than a sampling artifact. However, under naïve frequency-based estimates, we immediately face two related problems when dealing with massive class imbalances.

\begin{enumerate}
    \item Rare classes become nearly undetectable.
    \item Ignoring rare classes has minimal effect on macro-performance.
\end{enumerate}

To see this concretely, suppose we compare English ($\ell = \text{en}$) with a rare language ($\ell = r$). Let the global priors be:

\[
P(\text{en}) = 0.40, \qquad P(r) = 10^{-6}.
\]

Assume for illustration that the likelihoods of a given text under the two languages are of comparable magnitude—for example:

\[
P(X \mid \text{en}) = 10^{-4}, \qquad P(X \mid r) = 10^{-2}.
\]

Even though the rare-language model assigns a \emph{hundred times higher} likelihood to the input text, the posterior still overwhelmingly favors English:

\[
P(X \mid \text{en}) P(\text{en}) = 10^{-4} \cdot 0.40 = 4 \times 10^{-5},
\]

\[
P(X \mid r) P(r) = 10^{-2} \cdot 10^{-6} = 10^{-8}.
\]

Thus,

\[
4\times 10^{-5} \gg 10^{-8},
\]

\noindent and the argmax will select English. The only way the
 likelihood evidence can compensate for a prior that is effectively zero is if the likelihood for more common languages is \textit{also} effectively zero (likely due to a script mismatch). Because language $r$ occurs less than a thousandth of the time, getting it completely wrong will barely show up as a rounding error in overall accuracy. 
 

\subsubsection{Attenuated Frequency?}
Due to the extreme distribution of true frequencies, one common mitigation strategy is to upsample rare classes and/or downsample common ones \citep{burchellOpenDatasetModel2023}. This can help prevent a low $P(\ell)$ from instantly overpowering the likelihood evidence. However, even with this, there is a double bind. On the one hand, the prior still favors classifying data without obvious hints under a dominant class. On the other hand, the model can rely on a few shallow features  and thus lack robustness when such features appear as noise in real data. These two issues have led recent studies to emphasize False Positive Rate (FPR) as a key metric.

\subsubsection{False Positives at Scale}
To contextualize this issue, we revisit the example of \citet{caswellLanguageIDWild2020}, who note that a LID classifier that upsamples $r$ enough to achieve 99\% precision and 99\% recall on a balanced data, could be expected to recover 9.9K out of 10K true positives. However, if this is applied at web scale over 100B documents with a 0.01\% FPR, the resulting 10M false positives would render the resulting corpus unusable. Thus, adjusting training data ratios and carefully choosing metrics can reduce the damage, but they do not solve the underlying issue. 

Naturally, one is tempted to ask where these false positives come from. While the obvious answers of high-resource and out-of-model cousins do indeed play a role, \citet{caswellLanguageIDWild2020} also document  various forms of noise that are difficult to \textit{thoroughly} account for at training time. While a full review is beyond the scope of this article, important classes include emoji, misrendered PDFs, non-Unicode fonts, chance n-gram overlaps between unrelated languages, `A N T S P E A K', and mixed-script boilerplate. In response to this, they, as well as later filtered crawls like Fineweb-2 \citep{penedoFineWeb2OnePipeline2025}, emphasize \textit{post-filtering} for document-level consistency and token-precision in minority languages. Once again, this mitigates the fallout, but does not solve the issue.

\subsection{Local Priors?}
Once we appreciate the magnitude of the problem of false positives at scale, the central question is no longer how well languages can be discriminated in the abstract, but rather how to avoid suggesting labels that are not even plausible candidates in a given context.  We must first remember that, even for widely spoken languages, global frequency is often irrelevant. For instance, English, Mandarin, and Russian each account for a substantial share of web text, but each is typically written in a distinct script. Observing the script instantly render demographic dominance of the others nearly meaningless, and Bayesian models like CLD2 often incorporate preliminary script classification for this very reason. 

In theory, similar effects should arise when incorporating metadata such as geography, social networks, or user feedback. These contextual signals are useful in general, but crucial for assigning usable prior probabilities to  local languages, as well as managing the sheer size of the label inventory. In practice, however, converting context into usable prior probabilities remains challenging. Although some recent work has explored the utility of geodata \citep{dunnGeographicallyInformedLanguageIdentification2024a}, current benchmarks make it difficult to compare such context-aware models. The challenge is not simply that contextual information is missing; rather, the logistical requirements are fundamentally incompatible with how multilingual datasets are generally collected, labeled, and shared today.

\subsubsection{Where Do Languages Live?}
\label{Where Do Languages Live}
Firstly, there is the question of whether tail languages are actually used enough that misclassification is a practical, rather than theoretical, issue. Recently, many languages that have traditionally been considered ``unwritten'', ranging from European varieties like Luxembourgish \citep{wagnerLuxembourgishFacebookLanguage2013} to the Zapotec languages of Mexico \citep{lillehaugenTweetingZapotecSocial2019}, \textit{are} increasingly written (and spoken) on social media platforms and messaging services.  Over the last few years, both and privacy regulation and contractual restrictions have curtailed research access, but there is reason to believe this trend continues.


\subsubsection{Dataset Difficulties}
Given the limited availability of user data, public LID datasets turn to sources that are easier to work with, like Wikipedia, the Bible, or newswire.\footnote{Although even these are increasingly the subject of copyright challenges} Here, providing metadata about sources, by saying for example that the passage comes from a Saint Lucian Bible, would often solve the task, so the source is not given. Even the full, document-length text would still be to easy, so the tendency is to simulate harder scenarios by decomposing data from these sources into short, freestanding snippets, as in FLORES-200 \citep{nllb_teamNoLanguageLeft2022} and GlotLID-C \citep{kargaranGlotLIDLanguageIdentification2023}.

These practices can lead to a non-trivial circularity problem. Creating reliable labels for languages beyond the top 100–200 demands annotators with broad linguistic coverage, including familiarity with lesser-known scripts and regional varieties, as exemplified by \citet{kreutzerQualityGlanceAudit2022}. Even when such expertise can be assembled, annotators typically lack access to contextual metadata for the reasons we have just discussed. In practice, this is mitigated by beginning with pre-compiled corpora, which cover precisely the kinds of data for which LID already works.

\subsubsection{Model Expectations}

Older per-class models like CLD2 allowed users to provide hints like the expected language(s), URL sources, top-level domain name, original encoding, and embedded language tags. In contrast, newer discriminative classifiers like fastText and CLD3 that are deeply embedded in industrial and academic preprocessing pipelines cannot easily take local context into account. While it is possible to restrict the output space of such models to a few candidate languages, this involves an opaque manipulation of the softmax layer that can lead to unstable behavior.

\subsubsection{Towards Context-Aware LID}
In light of these operational difficulties, it is entirely understandable that public benchmarks do not focus on the relationship between the idealized task and real-world deployments. However, this encourages the development of models optimized for an version of LID that simultaneously overestimates the difficulty posed by rare languages in their local contexts and underestimates the impact of noise at scale. This makes it difficult to justify context-aware methods that target certain kinds of confusion that mainly surface when raw data is treated across diverse global sources. In Section \ref{Why  Is This Hard To Change?}, we will argue that this technical misalignment is reinforced by powerful institutional incentives, which further entrench the decontextualized framing of LID. However, we first present two case studies in Section \ref{Case Studies} which help illustrate the problem.


\section{Case Studies}
\label{Case Studies}
As a brief demonstration, we briefly consider two contact languages where the global prior probability is effectively zero, resulting in exclusion from all SOTA models. The first is Louisiana Creole (LC), a critically endangered language used online in the context of language revitalization. The second is the now-extinct Lingua Franca, long suspected to have influenced contact languages around the world.

\subsection{Language Revitalization}
\label{Language Revitalization}
Following over a century of language shift to English, LC is spoken today by a few thousand people. However, there is a revitalization movement with a strong online presence on several social media platforms, such as Facebook and Discord. In revitalization spaces, the main languages are
 LC and English. Some groups tolerate discussion of potentially relevant languages (Louisiana) French, Spanish, and other French-based Creoles, while others, like the Louisiana Creole Virtual Classroom prefer a strictly bilingual learning environment \citep{mayeuxLanguageRevitalizationSocial2024}. In any case, $P(LC)$ is much higher in dedicated groups than its global rarity would suggest.    

However, as the LID label set for existing models is defined at training time, the modeling assumption is that $P(LC)=0$. When downstream services like auto-translate are activated, the language is frequently misrecognized as French, which predictably leads to misunderstandings. Some platforms now have options to rate translations, guide the machine translation to a slightly closer target (i.e. Haitian Creole) or disable auto-translation.  However, at the time of writing, there is no user-facing option to identify a sentence as LC, reflecting the absence of LC from the underlying LID label set. As a result, users can mitigate downstream effects but cannot directly correct the initial language identification.

\begin{figure}
    \centering
    \includegraphics[width=\linewidth]{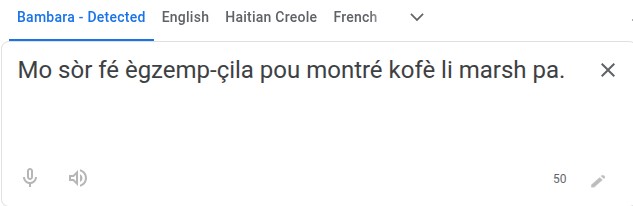}
    \caption{Interfaces allow override, but not for LC. (Google Translate)}
    \label{fig:google_translate}
\end{figure}

To explore this issue, we present a sample from Google Translate in Figure \ref{fig:google_translate} that is functionally similar to social media, but deliberately not taken from a real group due to privacy considerations. The input text is \textit{Mo sòr fé ègzemp-çila pou montré kofè li marsh pa}\footnote{`I just made this example to show why it does not work'}. The language is misidentified as Bambara and incorrectly translated as `I'm sure I did example to show coffee li marsh pa'. 

 What is of interest here is not the mistranslation, which is expected since the language is not supported, but rather the LID interface.  Previous work by \citep{yangItNavajoAccurate2025} has taken the absence of Native American languages as labels in Google Translate (and similar industrial models) as an indication that LID is a bottleneck. Yet, there \textit{is} an option for manual override. Crucially, this only extends to fully supported languages like Haitian Creole and French. Moreover, we are able to create a shortlist of  languages that have recently been used, and are thus presumed to appear again. Although the shortlist does not seem to bias the LID classifier, this interface showcases that for common languages, manual override is accepted and does not require a 7,000 language menu every time. Once we observe this capacity for manual selection, it becomes clearer that the question is not necessarily whether languages can be recognized, but rather whether recognition can be decoupled from full support. We return to this issue in Sections \ref{Commercial Incentives} and \ref{Transparency}.

\subsection{Retracing Language Contact}
\citet{whinnomSpanishLanguageContact1956} once argued that the Mediterannean Lingua Franca (LF) may have been an inspiration for contact varieties in Africa and Asia. Although the general idea has been largely rejected since the 1980s, \citet{nolanLinguaFrancaFrancais2023} claims LF may have influenced the 19th-20th century French military register known as Français Tirailleur (FT).  However, it has been extremely difficult to validate even this limited claim two main reasons. Firstly, LF was a rarely-written low status contact language based on --- yet distinct from --- prestige Romance languages like Italian and Spanish. Secondly, it was mostly used in North Africa and the Levant, where local writing was typically in the Arabic script and furthermore has not necessarily been digitized to the same extent.

Overall, current LID systems are not optimized for recovering  historical contact varieties as heterogeneous as LF, and existing models do not attempt to. Even if they did, the small extant corpus and high lexical overlap with much better documented Romance varieties as well as Romanized Arabic would likely lead to confusion with these more frequent categories. 

However, 
we have found reports which show that the French military served as a powerful vector for the global transmission of LF, also known as Sabir, following the French occupation of Algiers. In particular, the \textit{Relation historique de la
campagne du Sénégal (1861)
: extrait du journal d'un
capitaine de tirailleurs
algériens}\footnote{Historical Relation of the Senegal Campaign (1861) : Extract from the Journal of a Captain of the Tirailleurs Algériens} \citep{duboisRelationHistoriqueCampagne1863} includes several passages of dialogue in the contact variety during an expedition in Senegal around the time FT is thought to have developed. For example, \textit{Toi courageux kif r'azaloun, toi parlar kif barougue, toi tornar M'Boro, toi fasir meillour Lackmar.}\footnote{You who are courageous like a gazelle and speak like a parrot, you go back to M'Boro, you will do better than Lackmar.} In this excerpt alone, the pronouns \textit{toi} `you' is French, the names of the animals are Arabic, the verbs are Mediterranean Romance, and M'Boro is the name of a town in Senegal.

In a decontextualized setting, this combination would look like a code-switching nightmare. Yet, in the context of the 19th French \textit{Armée d'Afrique} `Army of [North] Africa' , the main task is to distinguish LF from French and to a much lesser extent, Arabic. This is a much more tractable sub-problem. In doing so, we identify a historically plausible mechanism of transmission in the form of Sabir-speaking Algerian troops transported to West Africa and placed in daily contact with newly conscripted Senegalese recruits. Although validating such historical connections is clearly feasible with lightweight NLP methods, such questions remain relatively marginal for reasons we explore in Section \ref{Why  Is This Hard To Change?}.


\section{Why Is This Hard To Change?}
\label{Why  Is This Hard To Change?}
Even if the limitations of decontextualized LID are clear, several distinct incentive structures across academia, industry, and public institutions prevent these problems from being addressed in practice.

 \subsection{Academic Incentives}
Within academic NLP, it is often  assumed  that meaningful contributions should involve novel architectures, training strategies, or datasets. Demonstrating that a well-understood technique still works is typically viewed as incremental rather than interesting. As a result, researchers are structurally disincentivized from relying on simple or well-established methods. This pattern is reflected in persistent under-citation of earlier foundational work \citep{singhForgottenKnowledgeExamining2023} and a broader tendency to treat re-engineering as innovation, had been amplified by the rapid growth of the field and the prestige hierarchy surrounding machine learning conferences. 

More specifically, \textit{Excitement} has been explicitly recognized as a key criterion for top-conference acceptance within the *ACL umbrella \citep{rogersProgramChairsReport2023}. This is well suited to promoting major breakthroughs, but it also deprioritizes contributions that are merely \textit{Sound}. Since LLMs are not yet a practical or cost-effective substitute for (shallow) general LID systems, there is an increasing gap between what the interests of the broader field and what actually works for this particular task.

To be clear, this dynamic does not strictly \textit{prevent} work like  \citet{vandergootIdentifyingOpenChallenges2025} from being published in main conferences. However, it does tend to route LID papers toward Findings and especially specialized workshops. The issue is not the quality of these venues, but their position in the prestige hierarchy. Because such venues lack the visibility and career incentives associated with the main ACL-branded conferences, there is little incentive for early-career researchers to specialize in LID, as opposed to other, ``more contemporary'' problems.



\subsection{Commercial Incentives}
\label{Commercial Incentives}

Equally, if not more important, is the reward structure of private industry. Here, the primary issue is not necessarily novelty or prestige. 
Rather, it is that existing LID solutions work well enough for commercially attractive languages, and improving LID for tail languages does not clearly lead to a viable revenue stream. Thus, even though there have been efforts such as \citet{caswellLanguageIDWild2020}, LID is not often treated as a pressing issue in its own right, but rather as a routing component for machine translation \citep[e.g.][]{nllb_teamNoLanguageLeft2022},  as explored in Section \ref{Case Studies}.

Furthermore, accurately identifying a language implicitly commits a platform to a certain level of support for that language. Once a system signals that it can reliably recognize a language, users may reasonably expect corresponding improvements in areas such as content moderation, abuse detection, policy enforcement, and ranking. However, fully supporting even one additional language often requires coordinated work across multiple teams such as policy, annotation, trust and safety, and regional compliance. This organizational effort can exceed the technical effort by a significant margin.

From a business perspective, the combination of increasing operational and compliance costs with no clear additional revenue is particularly discouraging. This in turn casts a large shadow on the data asymmetry problem, since, as noted in Sections \ref{Where Do Languages Live} and \ref{Language Revitalization}, industry platforms often contain precisely the high-quality, contextual data for rare languages necessary to build and evaluate robust LID systems.

\subsection{(Absence of) Vertical Integration}
 In principle, progress in ASR, transcription, text processing, and LID should be mutually reinforcing. ASR systems require textual resources; text collection is far more efficient when ASR exists; and accurate LID is a prerequisite for building both. This feedback loop operates naturally for many large languages.

For local languages, however, the previously named incentives lead to  structural deadlock. Each component waits for progress in the others, and only the largest actors have incentive to coordinate the pipeline. For example, some government and national security
agencies manage to overcome these issues and build vertically integrated systems, they cannot release them. It is true that some commercial endeavors like \citet{omnilingualasrteamOmnilingualASROpenSource2025} are starting to pay more attention to these interdependencies, but they have yet extend them to textual LID. Thus, even though public radio archives and contemporary broadcasts in principle provide enough material to put many languages into medium-resource territory \citep{doumbouyaUsingRadioArchives2021, havardSelfSupervisedModelsSpeech2025}, a lack of vertical integrations renders many of these resources effectively unusable for the moment.



\section{What Can We Do?}
In Section \ref{Context is Non-Negotiable}, we explored how using invariant priors poses special problems for languages that are globally rare but locally common. In Section \ref{Case Studies} we saw that such languages present particular challenges across social media and archives that are not insurmountable at the local level. In Section \ref{Why  Is This Hard To Change?}, we explored some institutional obstacles to incorporating contextual cues in LID models.  Now, we pivot from describing problems to outlining alternative solutions for these interwoven issues.

\subsection{Two Coherent Paths Forward}
\label{Two Coherent Paths}
In introductory computer science, we are explicitly taught to place correctness at the center of program design. A large part of correctness concerns pre-empting edge cases and testing them.
In LID, we recognize that most languages are edge cases, and simply eliminate them the label space during training. This is often justified as a necessity caused by data and/or model limitations. But is doing so correct?

The answer depends on one's conception of LID. If one accepts, as we have observed, that LID typically functions to route content within a larger system, then success consists of reliably moving content from Source A to Target B. This framing naturally treats LID as an engineering problem where solutions are evaluated relative to downstream requirements. If, however, we wish to maintain that LID remains a scientific problem, then it is imperative that the determination of prior plausibility be treated as problem in its own right. 

To make the role of prior specification concrete, we briefly sketch how local context can be incorporated into a standard per-class language identification model without introducing new architectures or training procedures. We define a \emph{local frame of reference} as any mechanism that restricts the hypothesis space of candidate languages or modifies their prior probabilities before observing the text itself, based on external signals such as document provenance, interface-level language settings, discourse continuity, previous inferences, or explicit user hints. Within such a frame, language identification consists of selecting the most plausible language relative to a locally valid prior distribution, rather than with respect to a single, globally fixed prior over all languages.

Concretely, let us (re)consider the Naive Bayes LID model represented in Section \ref{Context is Non-Negotiable}. 
Under a local frame of reference, the global prior, which implicitly encodes assumptions about worldwide language frequency, is replaced by a context-dependent distribution $P_{\text{local}}(\ell)$, yielding
\[
\hat{\ell} = \arg\max_{\ell \in \mathcal{L}_{\text{local}}} \left[ \log P(X \mid \ell) + \log P_{\text{local}}(\ell) \right].
\]

In the simplest case, this local context can act as a hard gate: languages deemed implausible in the current setting are removed from the hypothesis space, while remaining languages receive a uniform local prior. More intricate formulations allow for soft gating, in which locally plausible languages are upweighted relative to a global prior, while others retain a small but non-zero probability mass. Importantly, both variants substitute local priors for global ones at inference time, and do not require retraining the underlying language models. For coarse metadata on most languages, including alternate names, LinguaMeta \citep{ritchieLinguaMetaUnifiedMetadata2024} provides a usable starting point. More localized data, such as placing Native Alaskan languages in Alaska specifically,  would likely help limit false positives \citep[cf.][]{yangVisibilitySurvivalGeneralizing2025}.


We recognize that discrimination between very similar varieties remains challenging, and that even creating likelihood models requires a certain amount of data. This approach is fully compatible with hierarchical and multi-label setups that leverage specific similarities and differences between  related languages for fine-grained classification.

\subsection{Evaluation and Benchmarks}
One implication of the fork outlined above is that current evaluation practices conflate two distinct questions: whether a system can discriminate among languages given a fixed hypothesis space, and whether that hypothesis space is itself appropriate for a given use case. Under both paths, this conflation obscures the source of error and limits meaningful comparison.

As mentioned in Section \ref{Two Coherent Paths}, if LID is treated primarily as an engineering problem, then evaluation should be aligned with deployment conditions. Rather than optimizing for global accuracy over a fixed label set, systems should be assessed relative to the routing decisions they are intended to support, including downstream error tolerance, latency constraints, and the availability of auxiliary signals. In this setting, comparisons across platforms or domains are not expected to be universal. Instead, success is defined locally, in terms of whether content is routed to an appropriate handler with acceptable cost.

If, by contrast, LID is to remain a scientific task, then evaluation must explicitly separate prior determination from text-based inference. Benchmarks that present decontextualized text snippets implicitly encode an often arbitrary prior of convenience over the label space based on the available data. A scientifically meaningful alternative would require benchmarks in which non-linguistic signals are exposed as inputs and evaluated in their own right. This would allow models to be compared on their ability to restrict the hypothesis space before observing the text, as well as on their performance once that restriction has been applied.

Crucially, such benchmarks need not rely on sensitive or proprietary user data. As demonstrated by deployed systems, coarse signals such as interface-level language selection, document provenance, discourse continuity, or deliberately abstracted metadata can already produce substantial reductions in uncertainty.  Making these signals explicit would not simplify the task; rather, it would clarify which aspects of LID difficulty arise from linguistic ambiguity and which arise from prior misspecification.

Under either interpretation, the prevailing practice of evaluating LID systems solely on context-free text classification conflates engineering convenience with scientific validity. The result is not merely suboptimal performance for long-tail languages, but an evaluation procedure that is unable to distinguish between failures of modeling and failures of task definition.

\subsection{Transparency and User Interaction}
\label{Transparency}
As observed in Section \ref{Why Is This Hard To Change?}, institutions have various commitments that can make it difficult or unnecessary to support every language uniformly. For web platforms in particular, successful LID without corresponding downstream support can lead to complicated operational and organizational situations. However, in many spaces where tail languages are actually used, it can appear that LID is inherently difficult, when in fact performance could be meaningfully improved through local context injection.

It is worth reiterating that the issue is not simply reduced accuracy for these languages. Rather, they are systematically eliminated from consideration altogether, often through implicit prior assumptions. This occurs despite the fact that mechanisms such as manual override or constrained language selection (potentially combined with verification) are already accepted for widely supported languages. As such, even when such compensatory mechanisms are not implemented, making the exclusion of a language explicit is important. Setting
$P(\ell)=0$ \textit{everywhere} is a strong modeling decision, and leaving it implicit can lead to confusion when tested on languages where identification is often trivial due to script, such as Cherokee \citep{egglestonBridgingGapLeveraging2025}.

\section{Conclusion}
Despite the perception that language identification is an easy task, the majority of the world's languages are not still not consistently identified in writing. Correctly identifying such languages is a key step for reducing digital disparities across speech communities. One of the root causes of this gap is the difficulty of defining the local probability of a language. To move forward, we must pay special attention to the social contexts these languages are used in, and reconsider institutional practices that center novelty and modeling convenience.

\section*{Limitations}
We recognize that empirical demonstrations may have helped to make some of our arguments more concrete. However, due to the very nature of SOTA discriminative classifiers, directly exploring the effects of dynamic priors would have taken us beyond the scope of this paper.

\section*{Ethical Considerations}
Whether or not one agrees with our position, it should be clear that there is a certain tension between accurate coverage of highly localized language on one hand and privacy considerations on the other. In this paper, we have argued that using metadata like especially community and discourse structure is likely necessary to overcome technical obstacles to identifying languages that are not covered by existing large-scale systems. This does not imply that LID \textit{must} be implemented even in communities that express explicit concerns about privacy and/or surveillance. However, in cases where such considerations are a relevant factor, they should be stated alongside other engineering constraints.

\section*{Acknowledgments}

\bibliography{custom}




\end{document}